# Topic Modeling and Progression of American Digital News Media During the Onset of the COVID-19 Pandemic

Xiangpeng Wan, *Student Member, IEEE*, Michael C. Lucic, *Student Member, IEEE*, Hakim Ghazzai, *Senior Member, IEEE*, and Yehia Massoud, *Fellow, IEEE*

*Abstract*—Currently, the world is in the midst of a severe global pandemic, which has affected all aspects of people's lives. As a result, there is a deluge of COVID-related digital media articles published in the United States, due to the disparate effects of the pandemic. This large volume of information is difficult to consume by the audience in a reasonable amount of time. In this paper, we develop a Natural Language Processing (NLP) pipeline that is capable of automatically distilling various digital articles into manageable pieces of information, while also modelling the progression topics discussed over time in order to aid readers in rapidly gaining holistic perspectives on pressing issues (i.e., the COVID-19 pandemic) from a diverse array of sources. We achieve these goals by first collecting a large corpus of COVID-related articles during the onset of the pandemic. After, we apply unsupervised and semi-supervised learning procedures to summarize articles, then cluster them based on their similarities using the community detection methods. Next, we identify the topic of each cluster of articles using the BART algorithm. Finally, we provide a detailed digital media analysis based on the NLP-pipeline outputs and show how the conversation surrounding COVID-19 evolved over time.

## I. INTRODUCTION

At the time of this writing, there have been approximately 12.5 million (and counting) cumulative COVID-19 cases around the world, leading the world to buzz with mass uncertainty. The pandemic has seemingly upended all aspects of social life, and has presented serious human health and macroeconomic concerns, as many states, countries, and territories have instituted restrictive measures, such as shelter-in-place, stay-at-home, and lockdown orders. In many areas, the restrictive measures have led to tentative containment successes, at the expense of mass layoffs, changes in people's lives, and physical/mental health costs. Many of the public health measures taken to combat the spread of COVID-19 are novel, partially due to the novelty of the virus' characteristics and the unknown long-term outcomes of the infected [1]. Due to high rate in which COVID-19 can be transmitted, the virus can quickly infect large swaths of people in the aftermath of long-term success in arresting the spread, leading to far worse outcomes in the long-run. One debate that persists in the

United States is the tradeoff between the economic and health costs of the measures taken to control the spread of the virus. This debate has intensified as the pandemic has progressed, in part due to the reporting of both the health and economic effects of the virus. In the last five to ten years, there has been growing distrust of the media and expert opinion, in part due to the propagation of misinformation on the web. The effects of fake and misleading information on the web have led to profound effects in the real world, and researchers claim that societies are not ready for more advanced and readily available disinformation techniques such as deepfakes [2], [3].

Ideally, when journalists report on current events, the aim is to provide objective, unbiased reporting of the issues at hand in an appealing way for the consumers of news media. Most journalists strive to achieve this ideal; since journalists are human, and the institutions they work for may have an overarching agenda for how the news should be presented to the reader. Explicit and implicit bias tends to creep into the reporting of current events. Political news tends to exhibit these biases the most prominently, since politics are an inherently decisive topic. Bias at the institutional level tends to manifest in the way the publication reports information – certain views and stories may be given a more prominent position and receive disproportionate reporting. At the individual level, bias may manifest in the language that constructing an article, the journalist's own worldview and experience, or the sources information to construct the article [4]. Since different media institutions will inherently have their own agendas and their own staff that all have unique personal bias, the way an event is framed in an article may differ between institutions.

A prominent (and highly relevant example) is the American media coverage surrounding the 2009 H1N1 Swine Flu pandemic. The 2009 H1N1 pandemic occurred between January 2009 to August 2010 [5]. Ideally, coverage of a public health issue such as a pandemic would be unbiased. In the United States, reporting of the pandemic had some political bias introduced, as any large public action tends to have political opinions based on policy implementation, leading to differences in the framing of the story. Consider the differences in two media institutions and their reporting of the 2009 H1N1 pandemic: The New York Times and Fox News. The New York Times was founded in 1851 with the aim of providing non-sensationalized accounts of events, and thought its history has earned distinctions for the quality of its reporting, leading the Times to have the distinction of

Xiangpeng Wan, Michael C. Lucic, Hakim Ghazzai, and Yehia Massoud are with the School of System and Enterprise, Stevens Institute of Technology, Hoboken, NJ, USA. (Emails: {xwan6, mlucic, hghazzai, ymassoud}@stevens.edu).

winning more Pulitzer Prizes than any other media institution, while never being a leader in overall print circulation. On the other hand, Fox News was founded in 1997 with the aim of appealing to a conservative television audience, expanding into online media as the Internet became more prevalent; this has led to Fox News becoming the most-watched cable news provider in the United States. Fox News has been accused of curating biased and sensationalized content by several of its contemporaries [6]. These differences are clear when comparing examples of how the two institutions reported on the H1N1 pandemic. At the onset of the pandemic in the US, the New York Times wrote: "American health officials declared a public health emergency on Sunday as 20 cases of swine flu were confirmed in this country, including eight in New York City" [7]. Meanwhile, Fox News framed the same event as: "Did government health officials "cry swine" when they sounded the alarm on what looked like a threatening new flu?" [8]. This example provides a clear insight into how bias may affect the presentation of a current event, which shapes the worldview of the consumers of the information as well.

To reduce institutional bias, a reader can do one of the following: 1) read articles from a wide variety of sources, and synthesize the reporting of similar stories into moderated "averages" of the separate views, much in the same vein as the thinking behind the "wisdom of crowds" in statistics. 2) Learn about the motivations behind the editorial boards running an organization, as their collective views will shape how they frame stories brought to them by their staff writers. Both of these tasks are very time-consuming. Reading a multiplicity of stories covering the same event from different perspective forces geometric growth in the amount of time needed to develop a moderated view, and learning/staying informed of the always changing editorial boards and ownership of firms is very costly time-wise. To reduce time, automated systems may provide promise in reducing the time needed to aggregate various sources into their base information, which would be a key first step in systems that aim to reduce bias in media consumed by readers. Therefore, in this paper, we propose a two-pronged approach to reduce the effort required by the reader to aggregate different perspectives in the media they consume.

In order to satisfy the first approach, i.e., provide a moderate, complete perspective for the reader, our initial solution would be to aggregate articles based on topic similarity from different sources. In many cases, however, there is a huge volume of articles covering the same story; it is impossible for a reader to read all of them in a reasonable amount of time. Therefore, it is worthwhile to develop an automated system to search, collect, filter, summarize, and cluster the articles. In order to facilitate such a system, Natural Language Processing (NLP) techniques must be utilized. NLP is used to process and understand textual data in various applications including, but not limited to predicting customers' feelings towards a certain service [9] or product [10], [11] or rumor detection on social networks [12]. The NLP technology is now emerging in many online services and applications such as chatbots and sentiment analysis in social media. Microsoft has recently started to phase out its American editorial contractors in favor of an AI-driven system [13].

In this paper, we propose a generic NLP pipeline that collects, filters, and summarizes articles from various media sources with the aim of modelling the evolution of their discussion over time by leveraging *topic modeling* on the summarized articles. In this study, the automated discovery of the general idea(s) presented in a collection of text is referred to as *topic modeling*. The aim of this model is to provide a tool for readers to quickly consume the essence of the stories covered by a large multiplicity of articles, while also eliminating biased information that are prominent in each institution's reporting by only maintaining most important ideas of the articles and hence, isolating authors' opinions. We demonstrate that the model on a collection of articles from various news sources related to the recent COVID-19 pandemic, as this is the hot subject that has experienced long term discussion. We analyze the behavior of media with respect of its evolution within this subject.

The developed NLP-based framework is composed essentially of three main steps as shown in Fig. 1. After aggregating the news sources by events, we employ text summarization algorithms to distill the articles to their essential information. Generally, there are two main types of summarization, extractive summarization and abstractive summarization. Extractive summarization selects meaningful subsets of the article as the summarization result, whereas abstractive summarization reproduces new text as the summarization result. Although abstractive summarization is more advanced and closer to human-like interpretation, the extractive summarization still yields effective results. In this paper, we focus on applying extractive summarization to our digital media data set. Two algorithms for extractive summarization method are adopted to our framework: the unsupervised Google's TextRank [14] and the supervised Facebook's Bidirectional and Auto-Regressive Transformers for text summarization and generation (BART) model [15]. We compared the performance of both algorithms and adopt the most appropriate for our data set for the rest of the study.

After summarizing each article, the next step is to cluster the summaries of the articles and regroup them into groups containing articles discussing similar topics. The clustering algorithms that we consider are the Louvain and the Leiden methods, which automatically cluster the graph by maximizing the modularity of the graph [16], [17]. Finally, we run the designed topic modeling methods with the aim of discovering the main "topic" of discussion occurring in the collection of documents of each cluster. In the context of this paper, the aim of extracting the topics from the summarized articles is to gain insight on what the media institution of interest is discussing in their articles.

With the topics determined from the aforementioned process, we can utilize the pipeline to determine the main stories reported over time by various media institutions to get a high-level look at how they decide to frame their slate of stories, quantifying the views of the editorial board of that institution. The contributions of this paper are summarized as follows:

- Develop an automated framework that allows the readers to have a clear overview about the events covered by the

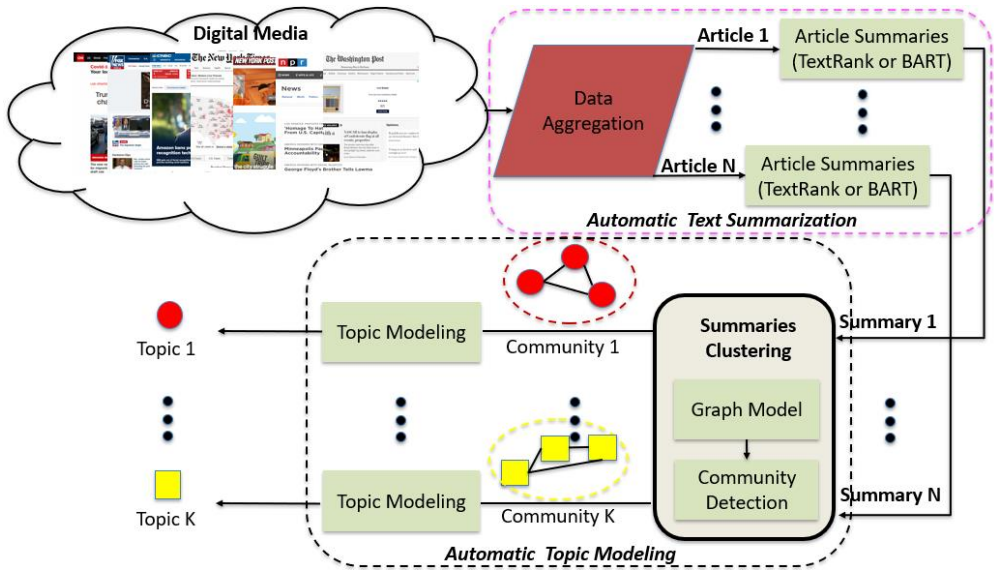

Fig. 1: Article analysis pipeline. Articles are aggregated and summarized. Afterward, they are clustered together based on similarity in chronological order, showing the evolution of discussion over time.

media with minimum influence.

- Design a series of NLP modules and graph analysis techniques to automatically cluster news articles into groups discussing similar topics while summarizing their contents.
- Apply the developed NLP framework to the case of articles on the COVID-19 pandemic and investigate the distribution of topics reported in various large American media sources.

The rest of the paper is organized as follows. Section II discusses selected related work. The proposed text summarization methods are presented in Section III. In Section IV, we introduce the topic modeling method and evaluate its performance. The results are discussed and concluded in Section V and Section VI.

## II. RELATED WORK

Interest in attempts to quantify media bias by way of leveraging state-of-the-art NLP methods has garnered much attention recently. Some studies explored novel methods for analyzing media bias. The authors of [18] curated a data set (named Bias Annotation Spans on the Informational Level, aka BASIL) and leveraged BERT to identify informational bias in articles based on the information captured in BASIL, where informational bias is defined as "sentences or clauses that convey information tangential, speculative, or as background to the main event in order to sway readers' opinions towards entities in the news." In other words, they worked to build a model that can identify bias that stems from the context in which information is presented. The authors of [19] explored the use of Matrix-based News Analysis (MNA) that attempts to quantify perspectives for the purpose of aggregating unbiased news. The authors of [20] considered two problems: the measurement of similarity between two articles, and whether or not they can match an article with its source based solely on its content. The second question also addresses bias, as matching content to a source would imply a form of institutional bias in the publication. The authors went about this by collecting articles from prominent polish online publications to curate a custom data set. To address each problem, they considered several machine learning methods to compare and select the most appropriate approach. They found that for both tasks, support vector machines provide the best results.

Other methods have focused on identifying bias based on the NLP pipelines. The authors of [21] explored how to quantify bias based on the selection of language to define a topic – they refer to this as Word Choice and Labelling (WCL). They present the example of how "economic mi- grants" versus "refugees" describe the same thing but evoke different emotional responses in the reader of the article. This bias is much harder to detect with computer systems. The main contribution of this article was the development of the NewsWCL50 open data set, along with presenting the basis for future research in this subject. The authors of [22] focused on the quantification of controversy in articles. They developed an application called "NewsPhi" that aims to present how controversial a topic presented in an article is to a reader. They utilize Latent Dirichlet Allocation (LDA) to build their controversy score model. They are motivated by the fact that more "controversial" content may stem from a large diverse set of opinions on a topic – by quantifying controversy the authors hope to present a tool that empowers readers to seek multiple perspectives on controversial topics as a way to combat bias.

Some researchers have employed some NLP features for some media related applications in the context of COVID-19. In [23], the authors proposed an NLP-based chatbot to answers the questions related to COVID-19 pandemic, the answers are extracted from reputable sources, to avoid misinformation spread from other unreliable sources. In [24], the authors proposed an unsupervised language model to identify false claims from different sources based on the perplexity scores. In this context, false claims are detected with high perplexity scores compared to true claims. In [25], the authors proposed a system that automated extraction the information about COVID-19 from social media and classified the text due to its sentiment using LSTM. In [26], the authors tracked and detected the COVID-19 ongoing events from tweets in Italy using NLP and graph analysis techniques. A term-frequency analysis is regularly performed to identify key words and understand the topics discussed at each time period. NLP is proven to be a powerful technology that can be used to devise several automated services and frameworks.

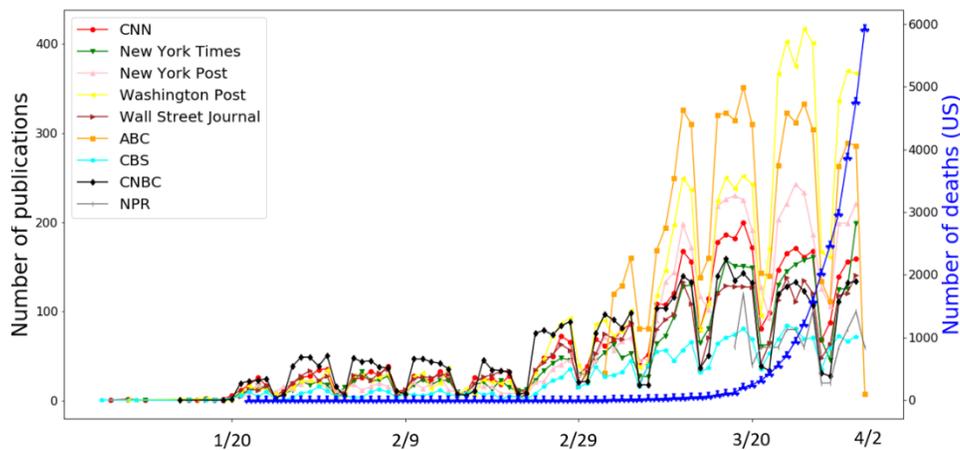

Fig. 2: The chronological evolution of the total number of publications related to COVID-19 in the US from selected media outlets, along with the number of American COVID-19 fatalities from January 1st, 2020 to April 2nd, 2020.

Our manuscript offers a unique contribution distinct from the related works by leveraging novel NLP- based summarization and clustering methods to present how the topics of discussion change over time. Unlike previous studies where one NLP technique is adopted to solve a single problem, in our framework, we design an automated framework composed of a pipeline of novel NLP models and graph analysis techniques to automatically analyze multiple articles from various media outlets and extract their main ideas while isolating authors' opinions and redundant details. This allow the readers to have a clear overview about the events covered by the media with minimum influence.

## III. AUTOMATIC TEXT SUMMARIZATION

In this section, we develop the text summarization part of our NLP pipeline. We compare the performance of the TextRank and BART methods on our data set generated by aggregating various COVID-related articles available on American digital media sites and select the model with the stronger general performance.

### A. Data Collection

As of the time of this writing, the COVID-19 pandemic has been a major event that has affected every aspect of the lives of Americans. The pandemic has led to (as of 30 July 2020) over 4.4 million confirmed cases and over 150,000 fatalities in just the United States (US) alone. In the US, the reporting of events relating to the COVID-19 pandemic has changed as the pandemic progressed from an outbreak in central China to the full-blown pandemic that is currently occurring. We aim to leverage our pipeline developed in this manuscript to gain insight into the change in discussion across various American media outlets. To this end, we gather 48765 articles related to COVID-19 from ten US media agencies: ABC, CBS, CNBC, CNN, The New York Post, The New York Times, NPR, and The Washington Post between 1 January 2020 and 4 April 2020. When scraping the articles, the program searches by three keywords: "COVID-19", "coronavirus", and "pandemic". The time frame of the articles collected roughly corresponds to the time frame between the first public reports of the COVID- 19 outbreak in Wuhan, China and the first peak of the COVID- 19 new confirmed cases in the Northeastern US. These media sources were selected because they are some of the largest and most influential media institutions in the US, and we chose the aforementioned time window as it contained the largest shift in US public discourse due to the rapid onset of effects of the pandemic. In Fig. 2, we visualize the number of COVID- related publications of each outlet of interest, along with the cumulative number of COVID-related fatalities, where the x- axis represents the date of publication and y-axis represents the number of publications/death count. We notice that over time, the number of articles published that related to COVID- 19 increased nearly in lockstep with the number of fatalities, with the seasonal behavior corresponding to the day of the week (the news cycle tends to slow down on weekends).

Along with our data set, we obtain a data set generated by the authors of [27], which consists of manually summarized articles from CNN and the DailyMail. In addition, the New York Times articles that we obtained contain subheaders that act as miniature summaries for the articles. These summaries act as the labelled training set for BART later on. The data set we collected contains nearly 50,000+ (and growing) articles over a three-month period. This is roughly 500+ articles per day if averaged uniformly across the time period, which is far too time-consuming for a typical reader to allocate their time to. As a result, the reader will tend to cherry-pick their sources, leading potentially to a biased or incomplete view of a subject. We address this through automated text summarization methods. Moving forward, we consider two summarization solutions: an unsupervised approach based on TextRank and a supervised approach that leverages the BART model.

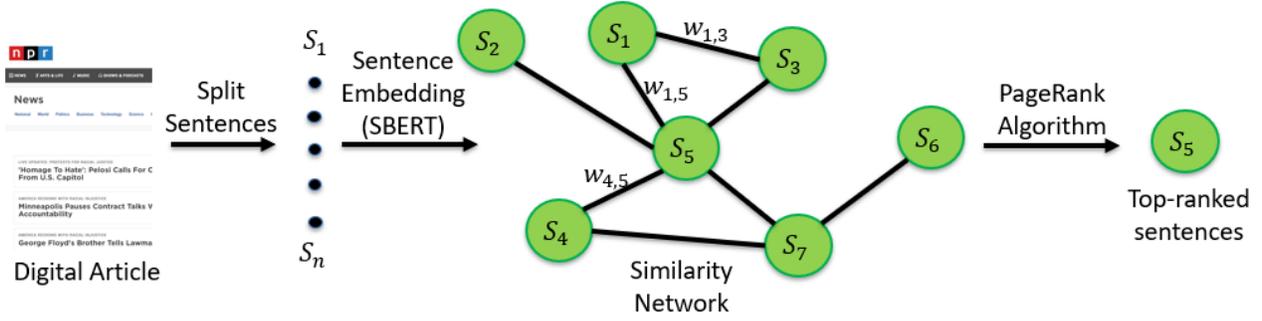

Fig. 3: Steps of the TextRank-powered approach for automatic text summarization. The sentences of each article are encoded by a word embedding method and reshaped into a semantic network by computing their similarities. Finally, the PageRank algorithm is executed to select the top-ranked sentences.

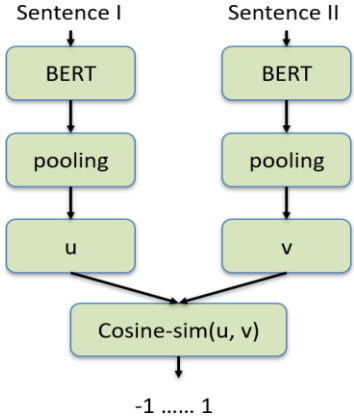

Fig. 4: High-level SBERT architecture with cosine-similarity objective function.

*B. Text Summaries from the TextRank Algorithm*

In this subsection, we develop the TextRank approach for our pipeline, which is visualized in Fig 3. First, we split the article into different sentences denoted by $S_i$ and create a similarity network. Each sentence is represented by a node in the network, and the edges connecting two nodes represent the similarity between the corresponding sentences that we associate to the edge weight denoted by $w_{j,j'}$. The similarity is calculated by cosine-similarity value between two vectors encoded from the two sentences. Traditional word-embedding algorithms such as Word2Vec [28] and GloVe [29] map words into vectors, while these early word embeddings were capable of capturing the context of the words in articles, there have been efforts to improve on the ability for embeddings to capture how varying context can alter the meanings of words in text. Google's Bidirectional Encoder Representations from Transformers (BERT) model is a state-of-the-art word- embedding that represents the next step in improving context modelling. BERT can consider the context of words adjacent to a word in either direction [30]. With the BERT encoder, each word would be converted to a 768× 1 vector, where each element of the vector exists on the range[ l, 1]. Its value not only depends on the word itself, but the context of the sentence. As an extension, an entire sentence of length $N$ is represented by a $768 \times N$ vector such that each element again exists on the [-1,1] range.

BERT has a drawback - it requires that the vectors utilized to calculate the cosine-similarity metric are of the same dimension. Researchers have utilized the average value of the BERT output layer as the encoded results, where each sentence vector is projected onto $768 \times N$, however this common practice leads to poor results. To address this, researchers have developed an extension of BERT for modelling sentences embeddings that named Sentence-BERT (SBERT), to encode the sentences [31]. SBERT uses Siamese and triplet networks that derive semantic sentence embedding with fixed output shape; the architecture of SBERT is shown in Fig. 4. The SBERT is pre-trained on SNLI [32] and the Multi-Genre NLI [33] with batch size of 16, Adam optimizer with learning rate $2 \times 10^5$, and a linear learning rate warm-up over 10% of the training data set. In short, SBERT achieves good performance on sentence embeddings compared to BERT in a computationally efficient manner, thus we decide to use SBERT in our work as our sentence encoding model.

The similarities between sentences calculated with SBERT allow us to assign the similarities to the edges so as to select the top-ranked sentences as the summary of this article by running the PageRank algorithm. The PageRank algorithm was utilized by Google in their original search engine product for measuring the importance of website pages based on the number of websites they linked to, along with the importance of the links contained [34]. The initial value of the edges between two nodes represents the probability that users may go through from one node to the other node. The values would be updated in an iterative fashion to get the final sentence ranking scores. The formula is shown as follows:

$$V(i) = (1 - d) + d \times \sum_{j \in In(i)} \frac{w_{j,i}}{\sum_{k \in Out(j)} w_{j,k}} V(j),$$

where V (i) is the score of the sentence i, $w_{j,i}$ is the weight of edges from node j to i, d is the damping factor in case of no outgoing links, In(i) is the set of inbound links from i, and Out(j) is the set of outbound links to j.

With the labelled data from the New York Times, CNN, and the DailyMail, we are able to evaluate the summarization algorithms by utilizing the Recall-Oriented Understudy for Gisting Evaluation (ROUGE) metric. It is common to employ the ROUGE metric in evaluating automated text summarization [15], [35], [36]. It allows to compare between the NLP generated summary against the manually written reference and calculates the overlap between them. There exists several ROUGE evaluation metrics. For instance, ROUGE-1 refers to the overlap of unigrams between the generated summary and the reference, ROUGE-2 refers to the overlap of bigrams, and ROUGE-L refers to longest common sub-sequence based F-measure, which finds the longest sub-sequence common in both sentences [37]. Note that sub-sequences are not required to be exactly matching permutations. The evaluation results are shown in Table I. Note that summarizing articles is a subjective task depending on many human factors, and therefore we cannot claim there exists an ideal human-written summary.

TABLE I: TextRank algorithm ROUGE-1,-2, and -L evaluation on the New York Times and CNN/DailyMail datasets

| Metric | R1 | R2 | RL |
|---|---|---|---|
| New York Times (LEAD-3) | 46.99 | 16.98 | 40.43 |
| New York Times | 47.77 | 14.70 | 44.83 |
| CNN/DailyMail (LEAD-3) | 40.42 | 17.62 | 36.67 |
| CNN/DailyMail | 42.60 | 17.19 | 36.70 |

We then compare the results of the TextRank summaries with the LEAD-3 baseline, which treats the first three sentences from an article as the summary. The LEAD-3 results for the validation data sets are also shown in Table I. Compared to the LEAD-3 baseline, the TextRank approach leads to slightly improved performance on both data sets. The TextRank approach has the additional benefit of being an unsupervised approach, so no manual labeling is required to generate automated summaries. Supervised learning approaches provide the potential for increased performance, at the cost of the requirement of manual labelling of the training data.

### C. Text Summaries from the BART Model

Generally, in most machine learning tasks, semi-supervised learning approaches tend to have more robust performance [38]. Over the past few years, there has been some exploration in the application of supervised learning to text summarization. These include: the PTGEN model [35], which achieves $R1 = 36.44$, $R2 = 15.66$, and $RL = 33.42$ on the CNN/DailyMail test set, the DRM model [39], which achieves $R1 = 41.16$, $R2 = 15.75$, and $RL = 39.08$ on the same data set, and the BERTSUMABS model [36], achieving $R1 = 41.72$, $R2 = 19.39$, and $RL = 38.76$. From these results, we notice a marked improvement in performance over unsupervised approaches, and therefore find it worthwhile to explore the application of state-of-the-art supervised methods on our data set. The current state-of-the-art in this area is Facebook's BART model [15], which is described by the authors as "generalizing BERT (due to the bidirectional encoder)". The model architecture is a modified variant of a transformer model [40], which utilizes the Gaussian Error Linear Unit (GELU) activation instead of the more-commonly used Rectified Linear Unit (ReLU) activation function. When applied to the CNN/DailyMail data set, the model achieves $R1 = 44.16$, $R2 = 21.28$, and $RL = 40.90$ [15], which outperforms other models on the same data set. Thus, we decide to explore BART model on our collected data set. The BART model we explore in this paper has 12 layers in both the encoder and decoder, for a total of 24 hidden layers. In order to apply BART to our data set, we first train it on 287227 articles from the CNN/DailyMail data set, then validate it on 11490 articles from the same data set along with 4776 new articles form the New York Times. The training period took 15 hours with an Nvidia V100 Tesla GPU – the results are shown in Table II.

TABLE II: BART model ROUGE-1, -2, and -L evaluation on the New York Times and CNN/DailyMail data sets

| Metric | R1 | R2 | RL |
|---|---|---|---|
| New York Times | 50.45 | 20.53 | 47.09 |
| CNN/DailyMail | 44.16 | 21.28 | 40.90 |

After determining the distributions for the BART and the TextRank models, we apply a statistical testing to the distributions of the ROUGE scores for both models, with the null hypothesis being that the means of the ROUGE scores are similar (i.e., mean of the ROUGE-X for BART minus mean ROUGE-X for TextRank where $X \in \{1, 2, L\}$), and the alternative hypothesis being that the means of them are different. For all three ROUGE scores, we compute the 95% confidence intervals (CIs) on the difference of the means of both distributions ($CI_{R1} = [0.0072, 0.0464]$, $CI_{R2} = [0.06, 0.08]$, $CI_{RL} = [0.003, 0.0422]$) and find that none of them contain the value 0, which means that the difference is statistically significant. We therefore reject the null hypothesis that the distributions are similar for all three ROUGE metrics. Since the difference of the means contain positive values, this implies that the BART ROUGE scores are superior. We note here that one pitfall of BART in practice is that sometimes it may hallucinate the content when summarizing text. Our choice of BART is largely justified on performance grounds, as all text summarization methods have their own quirks in practice. Following this, in the next section, we still proceed with both the BART and TextRank models in our pipeline for comparison purposes.

## IV. AUTOMATIC TOPIC MODELING

After obtaining the extractive summaries for the articles in our data set, the next step is to cluster the articles such that they are related by a common topic. In this section, we adapt a topic modeling algorithm to serve this purpose. As with the text summarization step in our pipeline, the proposed topic modeling algorithms require the text to be structured into matrices in order to be abstracted as graph structures. In this step, the vertices of the graph represent the summary from each article, and the edges represent the cosine-similarity between the article summaries (calculated with SBERT). We again set a threshold of 0.6 for the cosine-similarity values, where edges that have a score less than this threshold are deleted from the graph. This threshold value is selected to maintain consistency with our approach outlined in the previous section.

In this network, the objective is to cluster the summaries of the articles together based on the similarity of their content, where each cluster represents a unique topic class. We therefore leverage Louvain's method for community detection, a greedy procedure that quickly clusters communities in large networks [16]. In this context, the clustering problem can be formulated as the maximization of the network modularity metric $Q$ as follows:

$$maxQ = \frac{1}{2m}\sum_{jj'}\left[\delta_{jj'} - \frac{k_j k_{j'}}{2m}\right]\delta(c_j, c_{j'})$$

where $k_j$ and $k_i$ are the sum of the weights of the edges attached to nodes j and j', respectively, m is the sum of the weights in the graph, $\delta$ is the Kronecker delta function, and $c_j$ and $c_{j'}$ are the communities of the nodes.

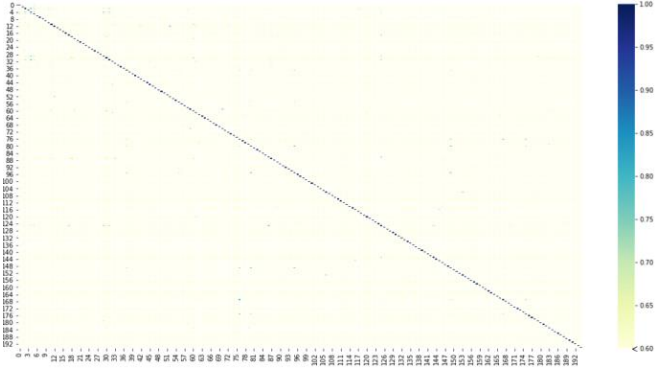

Fig. 5: Similarity matrix for the 195 New York times topics. From the plot, we notice that the clustered topics are highly unrelated, showing a successful clustering process.

TABLE III: Topic modeling results driven by Louvain's method for community detection, for only the first six topics

|  | Topic 1 | Topic 2 | Topic 3 | Topic 4 | Topic 5 | Topic 6 |
|---|---|---|---|---|---|---|
| Topic 1 | 1 | 0.27 | 0.51 | 0.47 | 0.22 | 0.17 |
| Topic 2 | 0.27 | 1 | 0.46 | 0.50 | 0.31 | 0.21 |
| Topic 3 | 0.51 | 0.46 | 1 | 0.53 | 0.58 | 0.32 |
| Topic 4 | 0.47 | 0.50 | 0.53 | 1 | 0.47 | 0.16 |
| Topic 5 | 0.22 | 0.31 | 0.58 | 0.47 | 1 | 0.38 |
| Topic 6 | 0.17 | 0.21 | 0.32 | 0.16 | 0.38 | 1 |

TABLE IV: Comparison between the community detection performances obtained using Louvain and Leiden methods from Mar. 26, 2020 to Apr. 2, 2020

| Criterion | Louvain | Leiden |
|---|---|---|
| Number of Communities | 101 | 104 |
| Largest Community | 1944 | 1820 |
| Average community size | 70.36 | 68.3 |
| Modularity | 0.6381 | 0.6377 |
| Coverage | 0.8177 | 0.8444 |
| Performance | 0.8097 | 0.7516 |

After applying the Louvain method to get a clustered set of topics, a community tag is assigned to different communities by running the BART model for the groups of sentences from each cluster. Consequently, the extracted community tags are the distilled topics that each article focused on during the period. To test the clustering approach, we first utilize 195 New York Times topics extracted from each cluster, calculate the similarities between them to check if the value is below our specified threshold of 0.6. Ideally, we would obtain a very small number of edges in the clustered network that are above the threshold, as this would mean that the articles cover sufficiently distinct topics. In the clustered graph, of the 18915 edges, only 124 (0.66%) of the edges are above this threshold; this is a sign that the clustering approach is working as expected. We show a subset of the numerical results (the first six topics) in Table III. Table V in the Appendix depicts the text of each corresponding topic (for the same six topics), where the New York Times focuses more on the international health and lack the vigilance of domestic cases. The full similarity matrix between the extracted topics of New York Times is visualized in Fig. 5. We notice that nearly all of the similarities between a pair of different clusters are less than 0.6, the threshold we select, which indicated that this is a good clustering. More examples of topics related to COVID-19 covered by the media between 1 January 2020 and 2 April 2020 identified by our proposed approach are provided in Table VI and Table VII in the Appendix.

On the other hand, some studies show that Leiden clustering algorithms provide more well-connected communities [17] than the Louvain approach does. The Leiden approach Achieves this by adding a refinement phase does not present the Louvain approach, potentially leading to improved quality of the partitions. With this consideration, we then apply Leiden clustering algorithms and the TextRank algorithm at the same stage of the pipeline that Louvain's approach is utilized to generate the topic summaries. The results of this are shown in Table VIII and Table IX in the Appendix, as compared to the BART generated result from Table VI and Table VII.

We also include the community detection performance of both the Louvain and Leiden methods in Table IV. Note that the performance of a partition is the ratio of the number of intra-community edges plus inter-community non-edges with the total number of potential edges and the coverage of a partition is the ratio of the number of intra-community edges to the total number of edges in the graph. We note that while the Leiden approach leads to better coverage of the topic clusters, the Louvain approach leads to slightly better performance with a slightly better modularity in terms of modularity. Statistically, both algorithms achieve almost the same number of communities with a very close average community size.

The developed framework presented in Fig. 1 is expected to work on large dynamic volumes of articles. It would likely depend on a distributed online implementation in practice, which cannot necessarily handle in devices equipped with low-power processors such as smartphones. The online training could occur in a remote location, which would then lead for the trained hyperparameters to be broadcast over-the-air to the user. Once the model is trained, applying new data to the trained model is quite trivial computationally. In this paper, we focus on training a static dataset in order to develop a basis on which an online system may be built upon.

## V. Analysis of Topic Trends

In this section, we apply our NLP pipeline to our data set of American media articles and discuss the reasoning behind the results we have found. We focus on the topics of discussion along all outlets across the selected time period, along with a focused analysis of the New York Times.

After clustering the article summaries from our data set based on similarity (and maintaining temporal consistency), we manually label the topics with commonly used media tags based on its distilled content (i.e., if the content includes information related to the American stock market, we label the topic under the "economics domestic" category). Examples of how these tags are applied to article summaries can be seen in Tables V, VI, and VII in the Appendix.

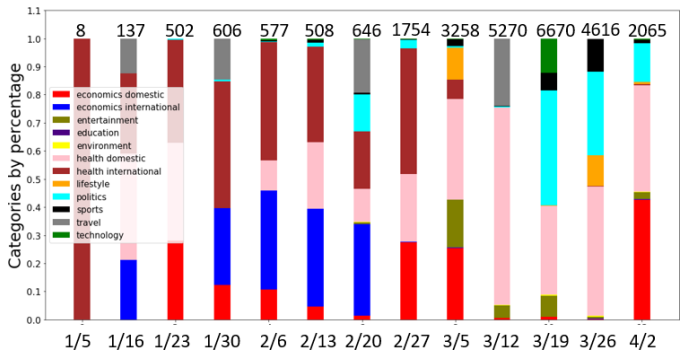

Fig. 6: The evolution of COVID-19 related topic discussion over time for all of our aggregated media sources. (The numbers on the top of each bar indicates the number of articles published during that week.)

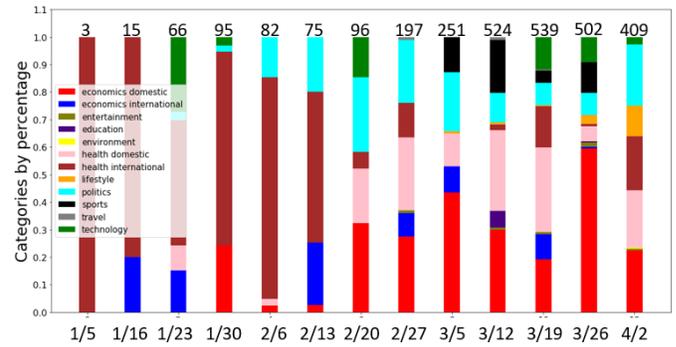

Fig. 7: The evolution of COVID-19 related topic discussion over time for the New York Times. (The numbers on the top of each bar indicates the number of articles published during that week.)

We then utilize these topic tags to get a general picture of the evolution of discussion in our collection of articles as the COVID-19 crisis proceeded. We count the frequency that a particular topic (and therefore high-level topic tag) is represented in a source at a particular date range (over a week). An example of how this is done is shown in the Table X in the Appendix. The topic evolution for our whole aggregated set of articles is visualized in Fig. 6, and the topic evolution for the New York Times is visualized in Fig. 7. We also consider the topics reported in all the publications of interest and visualize how each outlet focused their coverage of various COVID-related topics in Fig. 8.

We notice that, for all of the articles, the number of COVID-related articles grew rapidly from the onset of the pandemic to the beginning of April. In addition, we observe that the general topics of discussion evolved from mostly international health-related articles (as the COVID outbreak originated in Wuhan, China), to discussions split between domestic health and international travel, health, and economics (as the virus spread out to Europe) early on. As the pandemic began to affect the United States in early February, more discussion related to the US economy cropped up, and there was more discussion regarding the international economy. Eventually, as the outbreak began to heavily affect the US in late March and early April, the coverage began to focus much more on domestic issues: the economy, American public health, and the political ramifications of the outbreak. We notice an uptick in entertainment-, travel- and sports-related articles in late February and early March, which correspond to the travel restrictions and event cancellations that occurred to combat the domestic spread of COVID. Toward the tail end of the scope of our analysis, the discussion is dominated by domestic health and the US economy, which were both heavily affected by the outbreak along with the lockdown measures that were put in place to respond.

The coverage offered by the New York Times deviated somewhat from the reporting trends of the aggregated media sources. From Fig. 7, we see that the majority of the articles in the New York Times focused on domestic/international health, the US and global economics, politics, sports, and technology. The New York Times' coverage can be split based on coverage before and after the week of 20 February 2020. This week was a significant point, because the first detected cases of community spread (i.e., unknown origin) of COVID-19 in the US were found in California at this point in time. Detection of community spread meant that now the outbreak had reached a stage in the US where it would be much more difficult for public health officials to contain a widespread outbreak. Accordingly, the Times' coverage related to COVID-19 grew significantly after this point, and also shifted from coverage of international health (due to outbreaks in Mainland China and Western Europe) to domestic topics (health, economy, politics) as the US was more heavily impacted, especially the US economy. This increased economic reporting may have been a result of the publication's proximity to Wall Street in Manhattan, which is widely recognized as the global financial services capital.

After the consideration of the topic evolution over time, we analyze the coverage of each outlet (New York Times, New York Post, CNN, Washington Post, CNBC, and ABC) allocated to the general categories that we manually assigned to the topics. Fig. 8 provides an illustration of the coverage breakdown as pie charts, where each sector represents a source's percentage of articles dedicated to a topic. From this collection of charts, we notice a few patterns in the coverage. In all six of our selected sources, domestic issues dominated the coverage, with the combination of domestic health, economics, and politics comprising at least two-thirds of all articles from each source. In fact, articles relating to domestic health comprise more than 40% of all of the articles from the New York Post, CNN, the Washington Post, and ABC. The coverage offered by CNN and the Washington Post had near-identical breakdowns of topics discussed. The New York Post and ABC had the lowest coverage of international topics. The New York Times and CNBC had the most financial coverage, with both outlets dedicating over a third of their coverage to financial matters – the Times had a larger focus on domestic economic matters while CNBC had a more even split between international and economic matters.

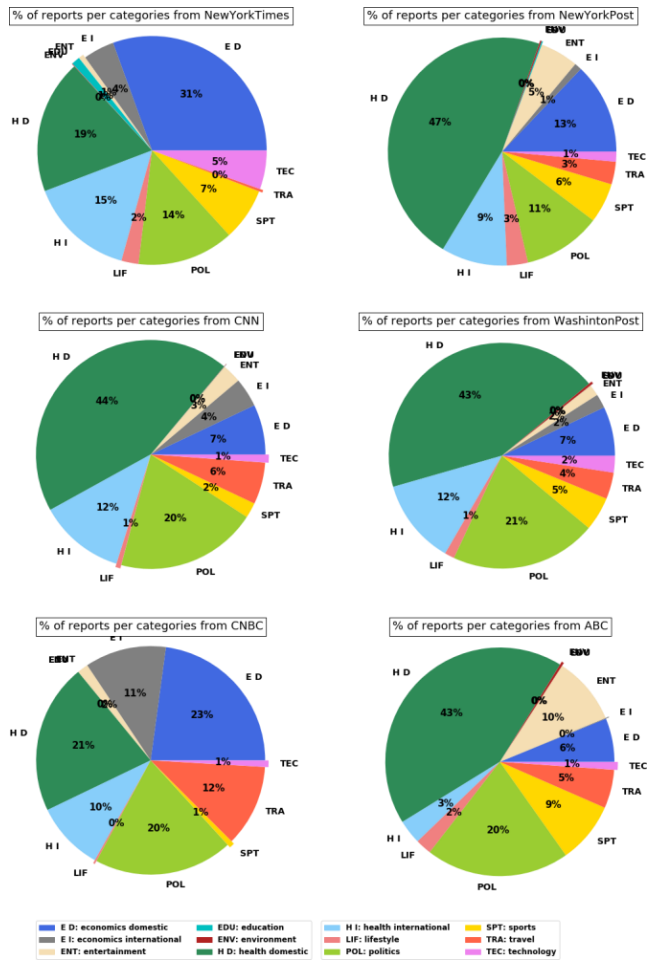

Fig. 8: Visualization of the coverage breakdown of the selected U.S. media sources based on the dominant identified category.

By segmenting the breakdown of topics reported by various agencies, those breakdowns may be compared to the breakdown of an aggregated mix of major sources. A significant deviation from the aggregation of sources could potentially indicate institutional bias in reporting and could further warrant more granular analysis of bias.

## VI. CONCLUSION

In this paper, we proposed and demonstrated an automated NLP pipeline designed to collect and summarize articles from a diverse set of sources, cluster them together based on the similarities of their core content, and visualize the breakdown and evolution of topics across various media sources. The user could visualize how different media focus on reporting on the same events by statistical report for stories and category evolution analysis. The model allows readers to identify key ideas of different articles and quickly follow the updates and evolution of different topics covered by media while eliminating redundant/unbiased information such as authors' opinions. Targeted bias detection remains a very challenging problem. The topic detection presented in this paper aims to be a first step toward the development of a tool to aid readers in getting the full, unbiased picture of current events. Such a system could work to incorporate our proposed process to distill information, along with some of the bias detection methods discussed in the literature review to measure how individual sources may deviate from the general topic discussed in the article.

Moving forward, we surmise a few key points of potential improvement: gathering a larger data set to explore other events and topics, building a classifier to automatically tag clusters to remove the only manual step currently in the pipeline, and exploring the information from social media and detecting misinformation by comparing with the reputable sources aggregation. In addition, it would be worthwhile to develop this system in a way to work in an online system to constantly update itself as new information arrives. As the world continues to output more information regarding current events, there exists a greater imperative to build systems to distill the deluge of information into succinct and informative memos.

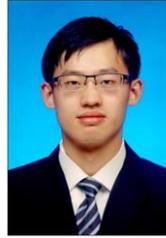

**Xiangpeng Wan** (S'12, M'15) is a Ph.D. student in Engineering Management at Stevens Institute of Technology, Hoboken, NJ, USA. In 2015, he received his Bachelor degree in Electrical Engineering from Harbin Institute of Technology, Harbin, China,and he obtained his Master degree in Applied Mathematics at University of Minnesota, Duluth, MN, USA, in 2017. His research interests are mainly on smart city design, big data analysis, and applied machine learning.

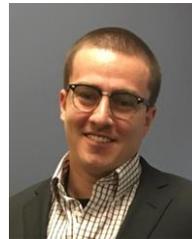

**Michael Lucic** (S'19) is currently a third-year Systems Engineering PhD student at Stevens Institute ofTechnology's School of Systems and Enterprises in Hoboken, NJ, USA. In 2018, he received a Bachelor of Science in Industrial and Systems Engineering with a minor in Mathematics from the University of Florida, Gainesville, FL, USA. His research interests revolve around Applied Optimization and Machine Learning in Intelligent Transportation Systems.

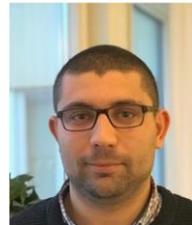

**HAKIM GHAZZAI** (Senior Member, IEEE) is currently working as a research scientist at Stevens Institute of Technology, Hoboken, NJ, USA. He received his PhD degree in Electrical Engineering from KAUST in Saudi Arabia in 2015. He received his Diplome d'Ingenieur and Master degree in telecommunications from the Ecole Superieure des Communications de Tunis (SUP'COM), Tunis, Tunisia in 2010 and 2011. Before joining Stevens, he worked as a visiting researcher Karlstad University, Sweden and as a research scientist at Qatar Mobility Innovations Center (QMIC), Doha, Qatar from 2015 to 2018. His general research interests include wireless networks, UAVs, Internet-of-things, and intelligent transportation systems.

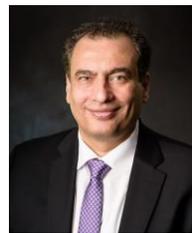

**Yehia Massoud** (Fellow Member, IEEE) received the Ph.D. degree from the Massachusetts Institute of Technology, Cambridge, USA. He is currently the Dean of the School of Systems and Enterprises, Stevens University of Science and Technology, USA. He is a Fellow of the IEEE and he has authored over 300 articles in peer-reviewed journals and conferences. He was selected as one of ten MIT Alumni Featured by MIT's Electrical Engineering and Computer Science department in 2012. He was a recipient of the Rising Star of Texas Medal, the National Science Foundation CAREER Award, the DAC Fellowship, the Synopsys Special Recognition Engineering Award, and two best paper awards. Dr. Massoud has held several academic and industrial positions, including a member of the technical staff with Synopsys, Inc., CA, USA, a tenured faculty with the Departments of Electrical and Computer Engineering and Computer Science, Rice University, Houston, USA, and the Head of the Department of Electrical and Computer Engineering, Worcester Polytechnic Institute, USA. Massoud has served as the editor of Mixed–Signal Letters–The Americas and also as an associate editor of IEEE Transactions on Very Large-Scale Integration Systems and IEEE Transactions on Circuits and Systems I

# APPENDIX

In the following, we provide five tables (Table V-Table IX) showcasing some randomly selected examples of the extracted topics summarizing some of articles published by the New York Times in different time periods. The first two tables are obtained using the NLP pipeline composed of BART and Louvain methods and the rest two tables are from anNLP pipeline consisting of TextRank/Leiden methods. The summaries colored in red correspond to the same stories generated by two different NLP pipelines. In Table X, we present some statistics of some of the topics discussed by different digital news media outlets.

TABLE V: Six examples of New York Times topics from 29 January 2020 to 5 February 2020 – note that these are the same six topics from Table III

| Categories | Topics |
|---|---|
| technology international | "Residents are using social networks and messaging platforms to offer on-the-ground accounts of the crisis that are difficult to find elsewhere. Chinese citizens are overcoming a lack of reporting on the crisis in the state-run media by sharing their own videos and information. Concern is growing that China's lockdown of cities may not only have come too late but could even make the situation worse." |
| health international | "A 44-year-old man who traveled from Wuhan, China, the center of the outbreak, died in the Philippines, officials said. If confirmed, it would be the first known instance of the virus in New York City. There are now eight confirmed cases in the United States and over 12,000 worldwide." |
| economics domestic | "U.S. stocks are poised to open down today, after Asian and European markets fell amid more signs that the Chinese coronavirus outbreak is continuing to worsen. American, Delta and United will halt service to the Chinese mainland, widening the impact of the outbreak on business and travel. The United States will beginfunneling all flights from China to just a few airports, including Kennedy International." |
| health international | "W.H.O. Declares Global Emergency as Wuhan Coronavirus Spreads. U.S. reports its first case of person-to-person transmission. China's Foreign Ministry spokeswoman said that the country "is fully confident and capable of winning the battle against this epidemic." |
| health domestic | "195 Americans who flew from China to California were first told they must clear medical tests that could take 72 hours or many days. Now they are all being quarantined for two weeks. Russia signaled that it was planning to close its 2,600-mile border with China amid fears about the coronavirus outbreak. Britain is scheduled to formally withdraw from the European Union on Friday." |
| politics domestic | "U.C. Berkeley Law School Drops Boalt Name Over Racist Legacy. The move to drop a name that has been used for decades. California Senate Bill 50: What's Next For Housing? Friday: A high-profile but divisive fix for the housing crisis is shut down." |

TABLE VI: Example of the extracted topics covered from Jan. 16, 2020 to Jan. 22, 2020 using the BART/Louvain methods

| Categories | Topics |
|---|---|
| health domestic | "More than 100 CDC staffers are being deployed to three US airports. They will check passengers arriving from Wuhan, China, for fever and other symptoms. Screenings at New York City's John F. Kennedy International Airport will start tonight. The monitoring comes as millions of people in China are traveling ahead of Lunar New Year. There are no known U.S. cases of the mysterious pneumonialike coronavirus." |
| health international | "Thailand has ordered thermoscanning of passengers in four Thai airports, including Suvarnabumi, Don Mueang, Phuket and Chiang Mai. The disease surveillance came about as a response to a severe pneumonia outbreak in Wuhan. Dozens of people have been hospitalized since mid-December – a handful in serious condition." |
| economics international | "Asian stocks recover from losses sparked by the coronavirus outbreak. China's Shanghai Composite was last up 0.4%, reversing earlier losses. Hong Kong's benchmark Hang Seng Index moved up 1.1%. South Korea's Kospi ( KOSPI ) was up1.2%, while Japan's Nikkei 225 (N225) was 0.6% higher." |
| health international | "Death toll from new coronavirus in China rises to six as new cases of the mysterious flu-like illness surged. Officials have confirmed more than 300 cases in China, mostly in the central city of Wuhan, where the virusmay have originated. Hundreds of millions of Chinese travelers are expected to travel both domestically and internationally as Lunar New Year starts Saturday." |
| health international | "A new Chinese coronavirus, a cousin of the SARS virus, has infected hundreds. Symptoms include a runny nose, sore throat, possibly a fever and maybe a headache. It's not clear how deadly the Wuhan coronavirus will be, but fatality rates are lower than both MERS and SARS." |
| health domestic | "U.S. officials on high alert as new cases of coronavirus spread in China. "Lean on Me" singer Bill Withers has died. Did a New York wedding lead to an attempted murder in Florida? Kids home from school face another viral threat: misinformation. Pentagon working to provide 100,000 body bags to FEMA." |
| travel domestic | "Travel suspended from Wuhan amid outbreak of deadly Chinese virus. Outbound flights were canceled Thursday from the central Chinese city. The death toll climbed to 17 on Wednesday with more than 540 confirmed cases ofthe flu-like illness. Moody's downgraded the city's credit rating to Aa3 from Aa2." |

TABLE VII: Some examples of the extracted topics covered from Mar. 26, 2020 to Apr. 2, 2020 using the BART/Louvain methods

| Categories | Topics |
|---|---|
| politics domestic | *"President Trump downplayed the number of ventilators he thinks New York will need in the coronavirus battle. New York Gov. Andrew Cuomo has requested 30,000 of the machines. Trump touted the trade deal he negotiated with China, but added that China has the "best dream in the world""* |
| health domestic | *"Number of coronavirus patients on ventilators doubles at NYC hospital. New York-Presbyterian/Columbia University Medical Center is relying on volunteer physicians to care for patients. More than 350 members of the NYPD were confirmed to have the virus Thursday. Prince Harry and Meghan Markle have fled coronavirus in Canada to set up a permanent home in California."* |
| health domestic | *"The Post is appealing to readers to donate protective gloves, surgical gowns and surgical masks to deliver needed protective equipment to public medical centers. Luxe labels are now using their posh factories to make masks for coronavirus. The number of confirmed cases in the US is the third highest in the world and dwarfs the number of cases in Mexico."* |
| lifestyle international | *"World-renowned Dutch flower garden Keukenhof in Lisse will not open this year. Dutch government extended its ban on gatherings to June 1 in an attempt to slow the spread of the coronavirus. A Van Gogh painting on loan to a small museum outside Amsterdam was announced as stolen on Monday."* |
| entertainment international | *"Berlin's nightclubs were closed March 13 to help slow the spread of the virus. In response, some of them formed a streaming platform to let DJs, musicians and artists continue performing. The shows run each night from different clubs between 7 p.m. and midnight."* |
| politics international | *"Thursday's virtual meeting of the Group of 20 nations, with more than a dozen heads of state participating, was less a global summit and more of a high-powered conference call. The meeting's purpose was to tackle the pandemic and its economic implications. The face-to-face tension among foes was gone."* |
| education domestic | *"Students at NYU's Tisch School of the Arts want tuition reimbursement. Their dean said no, offering a dance video of herself instead. Students pay $29,276 in tuition and fees, not including books or housing. More than 3,000 have signed an online Change.org petition.""* |
| health international | *"Turkey has so far reported 75 deaths related to the new coronavirus and 3,629 infections. Thousands of migrants had been waiting at the border with Greece hoping to make their way into Europe. Violent clashes erupted between the migrants and Greek border authorities trying to push them back."* |
| health domestic | *"The main list of acute symptoms at this time is actually quite short and can appear anywhere from two to 14 days after exposure to the virus. "We're emphasizing fever plus a notable lower respiratory tract symptom – cough or trouble breathing," said infectious disease expert Dr. William Schaffner."* |

TABLE VIII: Example of the extracted topics covered from Jan. 16, 2020 to Jan. 22, 2020 using the TextRank/Leiden methods

| Categories | Topics |
|---|---|
| health domestic | "More than 100 CDC staffers are being deployed to three US airports. They will check passengers arriving from Wuhan, China, for fever and other symptoms. Screenings at New York City's John F. Kennedy International Airport will start tonight. The monitoring comes as millions of people in China are traveling ahead of Lunar New Year. There are no known U.S. cases of the mysterious pneumonia like coronavirus." |
| health international | "Travelers from central China to LAX, SFO and JFK airports to be screened for new SARS-like illness. The monitoring — the first since Ebola — comes as millions of people in China are traveling ahead of Lunar New Year. There are no known U.S. cases of the mysterious, pneumonia like coronavirus." |
| economics domestic | "I'm betting the coronavirus will be the kind of exogenous event that lets you buy unrelated stocks at a discount," CNBCs' Jim Cramer said. The virus is "bad news" for travel stocks as the Chinese New Year nears. "In other words, if the whole market sells off tomorrow, that might be your chance to pick up some high-quality stocks into weakness," he said." |
| health international | "Coronavirus case discovered in U.S. as China death toll rises to 9" |
| health international | "Dow and broader stock market closed in the red Tuesday as reports of the first case of the Wuhan coronavirus in the United States weighed on the market. The Center for Disease Control and Prevention announced Tuesday that the first cases of the illness in the US showed up in Washington state. The virus, which was first identified in China, has infected more than 300 people in China and other Asian countries." |
| economics international | "Hong Kong stocks recorded their worst day in more than five months after renewed concerns about how the city is handling ongoing protests. Moody's downgraded the city's credit rating to Aa3 from Aa2. China confirmed that the Wuhan coronavirus — a disease that has killed at least four people and sickened more than 200 in the country." |
| economic international | "Mainland Chinese stocks made a turnaround to close higher after dropping more than 1% in the morning. The Bank of Korea said Wednesday the country's economy grew 1.2% on a seasonally adjusted basis in the fourth quarter as compared with three months earlier. Overnight on Wall Street, stocks declined after the Centers for Disease Control confirmed the first U.S. case of a mysterious coronavirus." |

TABLE IX: Some examples of the extracted topics covered from Mar. 26, 2020 to Apr. 2, 2020 using the TextRank/Leiden methods

| Categories | Topics |
|---|---|
| health domestic | "*New York Governor Andrew Cuomo praises first responders and says the coronavirus pandemic will define an entire generation of Americans. Record 6.6 million Americans file jobless claims Prince Charles gives message after Covid-19 diagnosis. College student makes masks for deaf and hard of hearing. Nurse struggled to gettested and worked while infected.*" |
| politics domestic | "*President Donald Trump made to accept reality of coronavirus in the face of America's growing death toll. Trump's decision to listen to his public advisers willrelieve public relieve and emergency physicians. Trump extends social distancing guidelines until April 30 Record 6.6 million Americans file jobless claims. Prince Charles gives message after Covid-19 diagnosis.*" |
| health domestic | "*Luxe labels are now using their posh factories to make masks for corona virus. The number of confirmed cases in the US is the third highest in the world and dwarfsthe number of cases in Mexico.*" |
| Lifestyle international | "*World-renowned Dutch flower garden Keukenhof in Lisse will not open this year. Dutch government extendedits ban on gatherings to June 1 in an attempt to slow the spread of the coronavirus. A Van Gogh painting on loan to a small museum outside Amsterdam was announcedas stolen on Monday.*" |
| entertainment international | "*Berlin's nightclubs were closed March 13 to help slow the spread of the virus. In response, some of them formeda streaming platform to let DJs, musicians and artists continue performing. The first livestream of the "United We Stream" project took place last week from the stageof Watergate.*" |
| politics inter-national | "*Thursday's virtual meeting of the Group of 20 nations, with more than a dozen heads of state participating,was less a global summit and more of a high-powered conference call. The face-to-face tension among foes was gone. The meeting's purpose was to tackle the pandemic and its economic implications.*" |
| education domestic | "*Tone-deaf NYU dean sends video of herself dancing to students seeking tuition refunds. Hundreds of studentsat NYU's Tisch School of the Arts demanded a partial tuition refund. Students say online classes and remote learning via Zoom are not worth the school's 58,000-a- year tuition.*" |
| health inter-national | "*Thousands of migrants had been waiting at the border with Greece hoping to make their way into Europe. Turkey has so far reported 75 deaths related to the new coronavirus and 3,629 infections. Greece hailed the development as an "important thing for our country and for Europe"* |
| health domestic | "*The main list of acute symptoms at this time is actually quite short and can appear anywhere from two to 14 days after exposure to the virus. "We're emphasizing fever plus a notable lower respiratory tract symptom – cough or trouble breathing," said infectious disease expert Dr. William Schaffner.*" |

TABLE X: Examples of statistical report for some topics reported during the investigated time period

| Topics | NYT | NYP | CNN | WTP | ABC | CBS | CNBC | Total number |
|---|---|---|---|---|---|---|---|---|
| "Since mid-December, 59 people have been diagnosed with viral pneumonia of 'unknown cause'. The new strain of coronavirus originated in Wuhan, the largest city in central China. It was confirmed Thursday to have been detected in Japan, a few days after Thailand confirmed its first case of infection." | 0 | 0 | 2 | 1 | 0 | 0 | 0 | 3 |
| "A new Chinese coronavirus, a cousin of the SARS virus, has infected hundreds. Symptoms include a runny nose, sore throat, possibly a fever and maybe a headache. It's not clear how deadly the Wuhan corona virus will be, but fatality rates are lower than both MERS and SARS." | 0 | 1 | 2 | 6 | 0 | 0 | 1 | 10 |
| "FDA warns Purell maker to stop claiming its hand sanitizers eliminate Ebola and MRSA. FDA doesn't allow brands to make claims about efficacy against viruses, such as flu. The agency said it was not aware of any hand sanitizers that have been tested against viruses." | 0 | 1 | 1 | 1 | 0 | 0 | 0 | 3 |
| "WHO declares China coronavirus a global health emergency. U.S. reports first person-to-person spread in spouse of Chicago woman. Nearly 200 Americans evacuated from Wuhan landed in Southern California Wednesday. The virus had killed 213 people — all of them in China — and infected more than 9,700 as of Friday" | 3 | 32 | 32 | 19 | 0 | 1 | 39 | 126 |
| "The risk of catching any serious viral infection during a flight is "very low" as the air on planes is purified with surgical-grade filters. The most crucial way to minimize exposure to the infection was to practice good hand hygiene. Coronavirus, which was renamed COVID-19 on Tuesday, is believed to be spread through respiratory droplets." | 0 | 1 | 0 | 0 | 0 | 0 | 2 | 3 |
| "Food And Drug Administration Seeks To Expand Treatment For Coronavirus : NPR. Food and Drug Administration. To expand treatment for Coronavirus. Founded in 1863, NPR is the world's first public radio station. The station's flagship program, Morning Edition, airs weekdays from 9 a.m. to noon ET." | 13 | 0 | 0 | 0 | 0 | 0 | 0 | 13 |
| "White House expected to urge Americans to wear face coverings in public to slow spread of coronavirus." I'm not going to wear a mask," says the dean of the Harvard Public Health School. The recommendations represent a major change in CDC guidance that healthy people don't need masks." | 0 | 37 | 7 | 48 | 5 | 0 | 10 | 107 |